\documentclass{article}

\PassOptionsToPackage{table}{xcolor}

\usepackage[blue]{b3}

\usepackage{amsmath,amsfonts,bm}

\def\eqref#1{equation~\ref{#1}}

\def\1{\bm{1}}

\DeclareMathAlphabet{\mathsfit}{\encodingdefault}{\sfdefault}{m}{sl}
\SetMathAlphabet{\mathsfit}{bold}{\encodingdefault}{\sfdefault}{bx}{n}

\usepackage{algorithm}
\usepackage{algpseudocode}
\usepackage{multirow}
\usepackage{tabularx}
\usepackage{array}

\definecolor{teleop}{HTML}{6A92CF}
\definecolor{rollout}{HTML}{77B383}
\definecolor{human}{HTML}{E8D171}
\definecolor{sim}{HTML}{D27858}
\definecolor{teleoptext}{HTML}{3E68A8}
\definecolor{rollouttext}{HTML}{3F8A55}
\definecolor{humantext}{HTML}{A88A1C}
\definecolor{simtext}{HTML}{B0502F}

\usepackage{tcolorbox}
\newtcolorbox{takeawaybox}{colback=teleop!8, colframe=teleoptext, boxrule=0.6pt, arc=6pt,
  left=5pt, right=5pt, top=1pt, bottom=1pt, before skip=4pt, after skip=6pt, fontupper=\small}
\newcommand{\takeaway}[1]{\begin{takeawaybox}\textbf{Takeaway.} #1\end{takeawaybox}}

\title{The Right Kind of Wrong: Counterfactual Data for Reward Learning in Manipulation}
\author{Gokul Kannan Thiruvenkadam, Vishal M. Patel\textsuperscript{\dag}, Homanga Bharadhwaj\textsuperscript{\dag}}
\jhuinstitution{\raisebox{-0.3\height}{\includegraphics[height=1.4em]{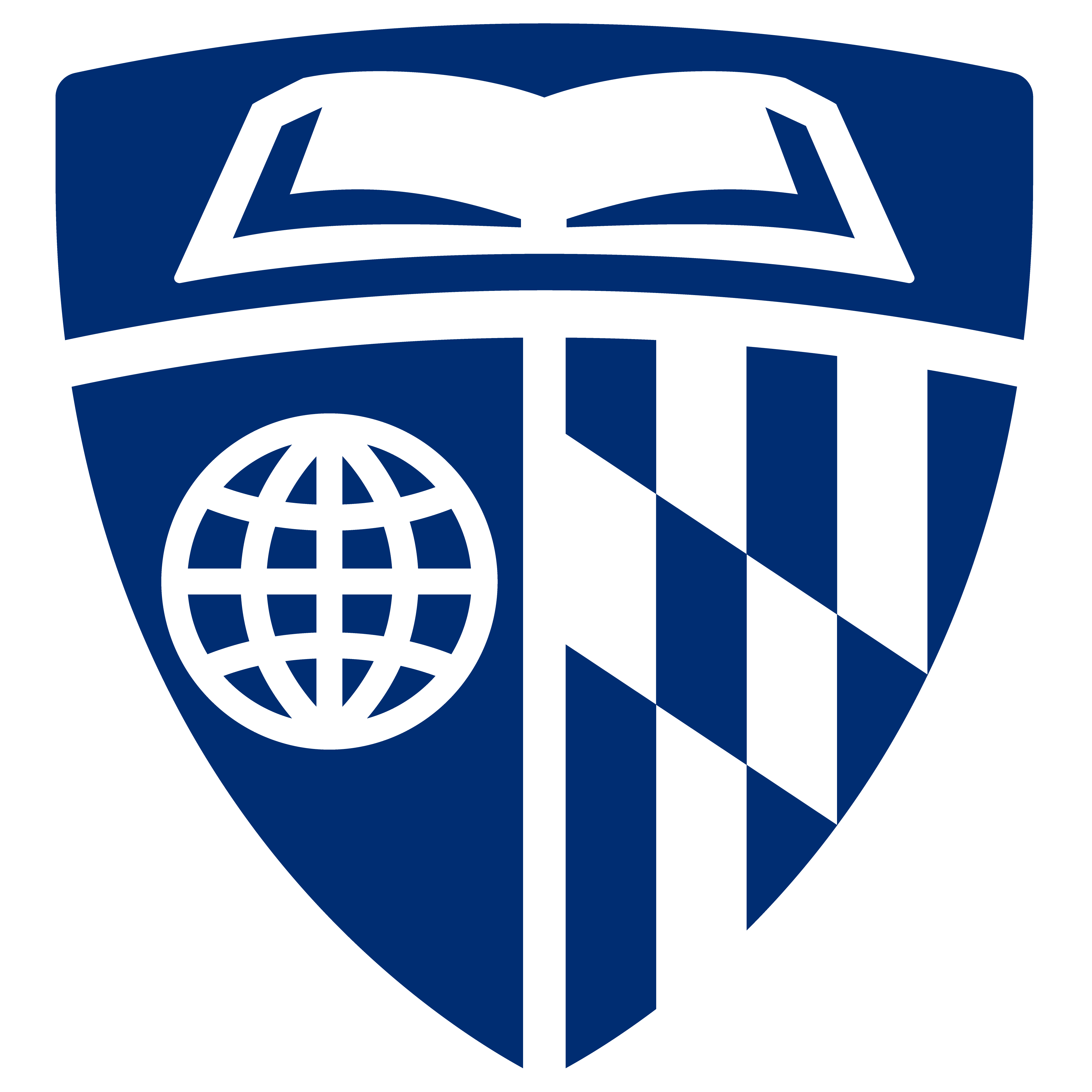}}\quad Johns Hopkins University \qquad \textsuperscript{\dag}Equal advising}
\jhucontact{%
  \raisebox{0pt}%
    [\dimexpr\height+8pt\relax]%
    [\dimexpr\depth+3pt\relax]%
    {\href{https://rightkindofwrong.github.io/}%
      {\texttt{rightkindofwrong.github.io}}}%
}
\begin{document}
\maketitle
\begin{figure}[H]
    \centering
    \vspace*{-0.5cm}\includegraphics[width=\linewidth]{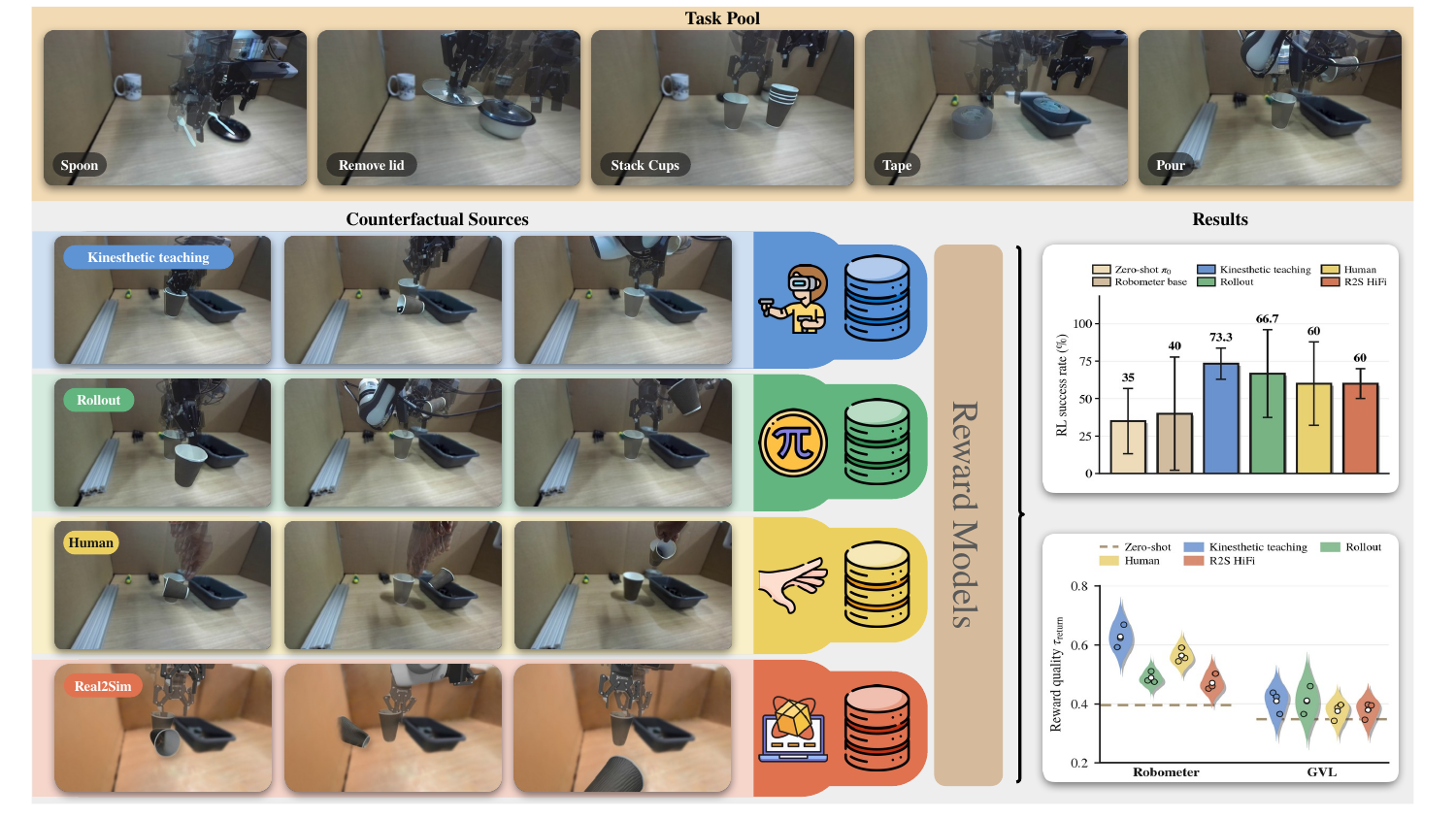}
    \vspace*{-0.8cm}\caption{\textbf{Overview of our analysis.} We consider counterfactual trajectories demonstrating failures obtained with kinesthetic teaching, pre-trained policy rollouts, human videos, and real2sim across five different manipulation tasks. We compare how the source of the counterfactuals and the scale affect learned reward models and subsequent robot-policy post-training. Video results are in the website \href{https://rightkindofwrong.github.io/}{\texttt{rightkindofwrong.github.io}}}
    \label{fig:teaser}
\end{figure}
\begin{abstract}
Reinforcement learning (RL) can improve robot policies beyond imitation learning, but it requires rewards that capture task progress and failure. Learning such rewards requires not only successful demonstrations, but also informative counterfactual failures that can guide the model on behaviors to penalize. However, large-scale robot datasets are dominated by successful behavior, and collecting failures on real hardware is expensive. This raises a practical question: given a limited data-collection budget, which sources of counterfactual experience are worth investing in? We compare four sources of failure data: kinesthetic-teaching failures, policy rollouts, human video, and real2sim rollouts, across five real-world manipulation tasks. By keeping the task suite, successful demonstrations, and the evaluation setup fixed, we study the effect of these sources through LoRA adaptation of a reward model and few-shot prompting of a frozen model in two different data scale scenarios. We evaluate the resulting reward models on a custom held-out benchmark of 250 human-annotated policy rollouts, measuring trajectory ranking and success detection. Finally, we use selected reward models to guide RL post-training of $\pi_0$ and test whether improvements in offline reward quality translate into better robot behavior. In sum, our study examines whether inexpensive counterfactual sources can match or surpass smaller amounts of faithful real-robot failure data when scaled, for improving rewards models in robot manipulation.
\end{abstract}

\section{Introduction}
Reinforcement learning (RL) can learn new behaviors from scratch that are hard to teleoperate \citep{yin2026emergent,zhang2026balanced} or improve large-scale robot policies trained via behavior cloning \citep{dong2026expo,wagenmaker2025steering,luo2025precise,lei2025rl}. Imitation-learned policies are a good starting point, but they still suffer from classic problems like compounding errors and covariate shift \citep{ross2010efficient,ross2011reduction,rajaraman2020toward}, and RL can help with these. Rewards are one of the main components for RL to perform well. Classically, there is a lot of engineering involved in shaping rewards \citep{gupta2022unpacking} to get better performance in RL. A well-calibrated reward needs to track progress for different varieties of rollouts, from successes to suboptimal ones to failures.

A growing body of work therefore studies learned reward models for robot manipulation. These models are typically trained on large-scale robot datasets \citep{o2024open, khazatsky2024droid}, which consist predominantly of expert demonstrations and contain little to no failure data \citep{atreya2025roboarena}. Existing methods thus need to obtain counterfactual supervision from several alternative sources. One category of works constructs synthetic counterfactuals from successful trajectories through augmentations such as reversing video segments, early stopping or stalling, and language relabeling \citep{zhang2025rewind, lee2026roboreward, torne2026freeform}. These can be viewed as a form of post-hoc relabeling to synthesize failures from successes. Another viable approach is to use real failures collected through deliberate teleoperation or policy rollouts on hardware \citep{liang2026robometer}. An alternative strategy is to avoid task-specific counterfactual data altogether, estimating rewards zero-shot from frozen vision-language models \citep{chen2026topreward, ma2025vision}. Recent work argues that reward models need bad behavior data to be reliable \citep{tian2026position}, but which kind is worth paying attention to? Synthetic failures are cheap but indirect, while real failures give a direct signal to learn from but are expensive.

This leaves a practical question that we set out to study: \textbf{What is the right kind of wrong for learning rewards?} We keep the tasks, reward models, training procedure, evaluation, and RL setup fixed, while varying the counterfactual data sources and how much data there is per source. Our most expensive sources are kinesthetic teaching (KT), where an operator physically guides the robot arm through a failure and the motion is replayed on camera, and zero-shot policy rollouts (RO), where we have no control over the behavior. The cheaper ones are deliberately collected failures in human video (HV), which creates an embodiment gap, and scripted failures in lifted real-to-sim environments (real2sim, R2S), which creates a visual gap. We first compare all sources at the same scale, and since the cheaper sources are easier to scale, we proportionally scale them according to their cost. 

As outlined in Figure~\ref{fig:teaser}, we study the counterfactual data sources across two reward models: one trainable, Robometer~\citep{liang2026robometer}, and one few-shot promptable method, GVL~\citep{ma2025vision}. We first test the adapted rewards on the real robot, using DSRL~\citep{wagenmaker2025steering} to post-train $\pi_0$~\citep{black2024pi0} on three tasks of increasing difficulty. 
Running RL for every source and condition we wish to study is not practical so we build a held-out benchmark of 250 annotated rollouts to study how source, scale, and the visual fidelity of simulation change the learned reward itself.  The cheaper sources surprisingly hold up well in RL despite their domain gaps, while more data and better offline scores do not always mean better reward signals.

\begin{tcolorbox}[colback=teleop!5, colframe=teleoptext, boxrule=0.6pt, arc=6pt,
  left=6pt, right=6pt, top=4pt, bottom=4pt,
  title=\textbf{Our main contributions}, fonttitle=\normalsize\bfseries,
  coltitle=white, colbacktitle=teleoptext]
\begin{itemize}
    \item \textbf{A controlled study of counterfactual sources.} We collect data from four sources across five real manipulation tasks and build a held-out, 250-episode cross-policy benchmark with human-annotated partial credits.

    \item \textbf{Cheap counterfactuals can also help real-robot RL.} With DSRL on three tasks, we find that every source improves over the base reward. Cheaper sources are competitive with the expensive ones despite their domain gaps.

    \item \textbf{What the offline benchmark shows.} Offline experiments show that the source matters more for trainable reward models than for prompted ones, that scaling helps only up to a point for cheap sources, and that offline metrics only loosely predict RL success.
\end{itemize}
\end{tcolorbox}

\begin{figure}[H]
    \centering
    \includegraphics[width=\linewidth]{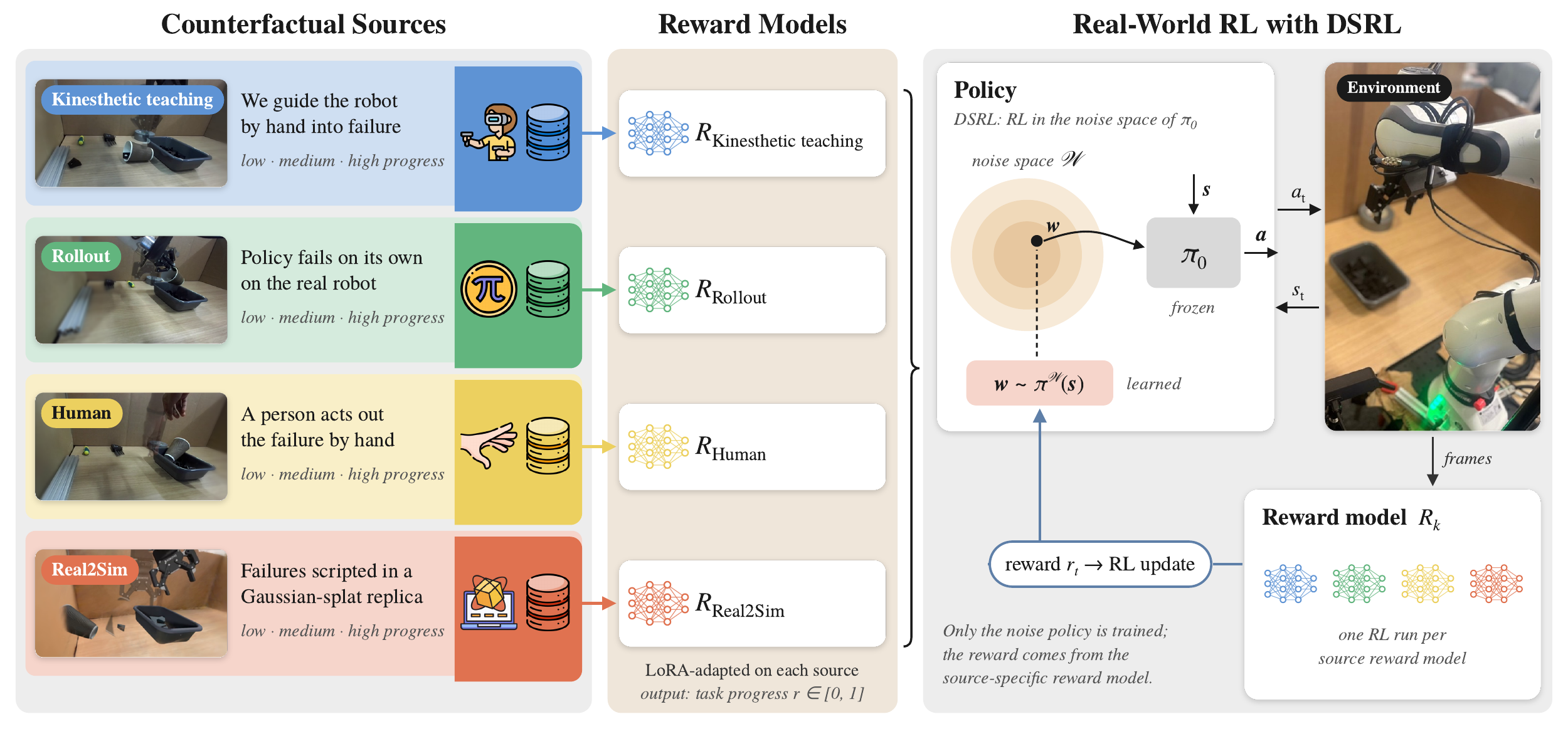}
    \caption{\textbf{Setup overview.} Left: each counterfactual source is used to adapt its own Robometer reward model with LoRA, which predicts task progress in $[0,1]$. Right: DSRL trains only a noise-space policy on top of the frozen $\pi_0$, and the reward for each RL run comes from one source-specific reward model.}
    \label{fig:setup}
\end{figure}

\section{Related Work}

\paragraph{Reward learning from demonstrations}
Learning rewards or value from demonstrations is a long-studied alternative to human-engineered rewards. Inverse reinforcement learning infers a reward that explains expert behavior \citep{abbeel2004apprenticeship, ziebart2008maximum}, ranking-based methods extrapolate beyond the demonstrator \citep{brown2019extrapolating}, and representations such as VIP \citep{ma2022vip} and LIV \citep{ma2023liv} measure progress as distance to a goal. These methods benefit from the growing supply of datasets dominated by expert behavior \citep{o2024open, khazatsky2024droid}, but successes alone cannot teach a reward to reliably score partial completions or failures. ReWiND \citep{zhang2025rewind} addresses this by reversing successful videos to synthesize undo behavior, a common failure seen in RL rollouts, which lets it learn decreasing progress. However, such augmentation-based failures do not cover the plausible failures a robot actually makes. We instead ask which sources of more realistic failures are worth collecting.

\paragraph{Reward learning from failures}
A second line of work supervises rewards with mixed sources of data. Recent methods like RoboReward~\citep{lee2026roboreward}, Robometer~\citep{liang2026robometer}, Robo-Dopamine~\citep{tan2025robo}, VLAC~\citep{zhai2025vision}, and LRM~\citep{wu2026large} train VLM-based reward or progress models with mixed data sources. We mainly use Robometer, which combines frame-level progress with trajectory-level preferences over successful, suboptimal, and failed rollouts across many embodiments. Failures also appear as first-class data in vision-language-action models~\citep{lin2025failsafe}, autonomous skill improvement~\citep{zhou2024autonomous}, and language-guided recovery~\citep{dai2025racer}, and a recent position paper~\citep{tian2026position} argues for the need for reliable bad behavior data. These works establish that failures are important; we study what kind of failures to invest in.

\paragraph{Rewards from foundation models}
A third line of work uses large pretrained vision-language models as general-purpose reward or value estimators. GVL~\citep{ma2025vision} treats a VLM as an in-context value learner that predicts progress over shuffled video frames, enabling zero-shot and few-shot progress estimation across tasks and embodiments. TOPReward~\citep{chen2026topreward} leverages a pretrained VLM's token probabilities to estimate task progress, whereas few-shot methods like RoboCLIP~\citep{sontakke2023roboclip} and RARM~\citep{yang2026rarm} score rollouts using their video-language similarity with the reference demonstration.
These approaches need no task-specific training, or at times no training at all, inheriting from broad visual priors, but their scoring depends on the reasoning capabilities of the foundation model they use \citep{bai2025qwen3,radford2021learning}. We include few-shot GVL as one of our two reward models to test whether counterfactuals still help when they are only seen in context.

\paragraph{Counterfactual data for manipulation.}
The sources we study correspond to naturally available data used in robot learning. Teleoperation and policy rollouts reflect large-scale datasets such as DROID~\citep{khazatsky2024droid} and Open X-Embodiment~\citep{o2024open},  RoboArena~\citep{atreya2025roboarena}, and collectively these sources show how expensive controlled hardware trials remain. Human video datasets like Epic-Kitchens~\citep{damen2020epic} and EgoDex~\citep{hoque2026egodex} have been used as cheap sources for robot learning across many tasks; even Robometer uses Epic-Kitchens in its training recipe. Simulation benchmarks such as LIBERO~\citep{liu2023libero} and RoboCasa~\citep{nasiriany2026robocasa365} provide cheap, exactly labeled rollouts at the cost of visual and dynamics gaps, which real-to-sim methods narrow by lifting the real scene into simulation~\citep{ranawaka2026simfoundry, chen2026agentic}. Prior works typically study or use these sources in isolation for policy learning, but we compare them for reward learning under a common cost model, annotation scheme, and evaluation protocol.

\section{Study Design}
\label{sec:design}
 
We design our study around the question: \emph{Given a limited data-collection budget, which sources of counterfactual data most improves learned reward models for reinforcement learning?} We keep the reward models, task suite, offline evaluation benchmark, RL setup fixed and vary only the \emph{source} of counterfactual data and the \emph{scale} collected from each source. 
 
\subsection{Robot setup and tasks}
\label{sec:tasks}
 
We evaluate on five manipulation primitives using a standard DROID~\citep{khazatsky2024droid} setup with a Franka Emika Panda arm and a Robotiq gripper as seen in Figure~\ref{fig:setup}. The tasks are: \emph{Pick up the spoon and put it on the plate} (Spoon), \emph{Pick up the sellotape and put it in the grey box} (Tape), \emph{Remove the lid from the pot} (Remove Lid), \emph{Pour the contents of the cup into the grey box} (Pour), and \emph{Stack up the cups together} (Stack Cups).

\subsection{Counterfactual sources}
\label{sec:sources}
 
We collect counterfactual data from four sources (Table~\ref{tab:sources}), each offering a different trade-off between collection effort and alignment to the reward-learning setting. Per-episode cost for every source is determined by the \emph{collection}, the time an operator spends generating a trajectory, and \emph{annotation}, the time a human spends assigning partial-credit labels. Sources that require the physical robot for collection are substantially more expensive per episode than ones that do not require it, and sources whose labels can be obtained automatically remove the annotation cost entirely. This naturally creates a clear cost ordering across sources. Source examples and the partial-credit labeling scheme are given in Appendix~\ref{app:counterfactual_sources}.
 
\begin{table}[t]
\caption{Counterfactual sources and their cost-alignment trade-off.
\emph{Collection} is the operator time to produce one trajectory;
\emph{annotation} is the time to assign partial-credit labels.
Robot successes are pooled into a shared set (Section~\ref{sec:rewards}).}
\label{tab:sources}
\centering
\footnotesize
\setlength{\tabcolsep}{4pt}
\renewcommand{\arraystretch}{1.3}
\begin{tabularx}{\linewidth}{@{}
  >{\raggedright\arraybackslash}l
  c c c c
  >{\raggedright\arraybackslash}X
  c
  >{\raggedright\arraybackslash}l
@{}}
\toprule
\textbf{Source} & \textbf{Embod.} & \textbf{Real vis.} & \textbf{On-policy} & \textbf{Scripted} &
\textbf{Per-episode cost} & \textbf{Annotation} & \textbf{Data Collected / task} \\
\midrule
\rowcolor{teleop!15}
\textcolor{teleoptext}{\textbf{Kinesthetic}} &
\checkmark & \checkmark & -- & \checkmark &
$t_{\mathrm{collect}} + t_{\mathrm{replay}}$ (kinesthetic + reset) &
human & 5 (4 fail\ + 1 succ) \\
\rowcolor{rollout!15}
\textcolor{rollouttext}{\textbf{Rollout}} &
\checkmark & \checkmark & \checkmark & -- &
$t_{\mathrm{rollout}}$ (policy on hardware) &
human & 20 $\pi_0$ rollouts \\
\rowcolor{human!18}
\textcolor{humantext}{\textbf{Human}} &
-- & \checkmark & -- & \checkmark &
$t_{\mathrm{human}}$ (no robot needed) &
human & 20 (19 fail\ + 1 succ) \\
\rowcolor{sim!15}
\textcolor{simtext}{\textbf{Real2Ssim}} &
\checkmark & -- & -- & \checkmark &
$t_{\mathrm{asset}}/N$ (amortized setup) &
automatic & 30( 29 fail\ + 1 succ) \\
\bottomrule
\end{tabularx}
\end{table}
 
\textbf{Kinesthetic-teaching failures} are the most expensive and most aligned source. An operator uses the Franka's program mode to kinesthetically record each trajectory by physically guiding the robot arm, then rearranges the objects to the exact same initial configuration and replays the recorded action.
Hence, each episode requires a phase of kinesthetic teaching and is followed by a physical replay phase, both of which demand direct robot access. Per task, we collect one expert trajectory and the rest as intentional failure modes, all recorded at 15--20\,Hz and annotated by a human with partial-credit labels. Per episode, the cost is $t_{\mathrm{collect}} + t_{\mathrm{replay}} + t_{\mathrm{annotate}}$, where collection and replay together make this the most time-consuming source.
 
\textbf{Policy-rollout failures} require the robot but not an operator guiding it. We deploy our base policy $\pi_0$-DROID~\citep{black2024pi0} on hardware and collect approximately 20 rollouts per task, recorded at 8--10\,Hz due to policy latency. Because the trajectories come from the policy rollouts, they are on-policy task behavior that the learner actually reaches, but we have no control over the distribution of successes and failures. Each episode requires a rollout on the robot ($t_{\mathrm{rollout}}$) followed by human annotation ($t_{\mathrm{annotate}}$). The collection cost per episode is roughly half that of kinesthetic-teaching data, since there is no kinesthetic teaching.
 
\textbf{Human-video failures} remove the robot from the loop entirely. A person performs the task by hand in the same scene, producing one expert trajectory and the rest as intentional failure modes per task. Collection is fast ($t_{\mathrm{human}}$): it requires no robot setup, no kinesthetic teaching, and no physical replay, making it substantially cheaper per episode than either robot-based source. However, the resulting data crosses an embodiment gap between human hands and the robot arm, reducing its alignment to the target setting. Trajectories are annotated with partial-credit labels by a human ($t_{\mathrm{annotate}}$). We collect 20 episodes per task for this source.
 
\textbf{Real2sim failures} have the lowest marginal cost per episode. We use components of real-to-sim methods such as SimFoundry~\citep{ranawaka2026simfoundry} and Agentic Real2Sim~\citep{chen2026agentic} to build articulated 3D assets from the real scene and arrange them in simulation. The effort for asset generation ($t_{\mathrm{asset}}$), which is a one-time upfront cost per scene.  Once the scene is built, failures are generated by scripted perturbations, and labels are assigned automatically from the simulator state, so both collection and annotation scale at near-zero marginal cost. Each task requires only one real-world successful demonstration (obtained from kinesthetic replay) to define the task; we then generate 29 failure trajectories and one replayed success per task. Real2sim scene construction and failure generation are detailed in Appendix~\ref{app:real2sim} and Algorithm~\ref{alg:failgen}.
 
Taken together, the per-episode cost ordering is intuitively:
\begin{equation*}
\underbrace{t_{\mathrm{collect}} + t_{\mathrm{replay}}}_{\text{KT}} \;>\; \underbrace{t_{\mathrm{rollout}}}_{\text{RO}} \;\gg\; \underbrace{t_{\mathrm{human}}}_{\text{HV}} \;>\; \underbrace{t_{\mathrm{asset}} / N}_{\text{R2S (amortized)}}
\end{equation*}
where all sources except real2sim additionally incur $t_{\mathrm{annotate}}$ for human labeling. This cost structure is what makes the scaling question meaningful, as the cheaper sources can get more data per unit of operator time and our experiments test whether that quantity compensates for their weaker alignment. 
\begin{figure}[!t]
    \centering
    \includegraphics[width=\linewidth]{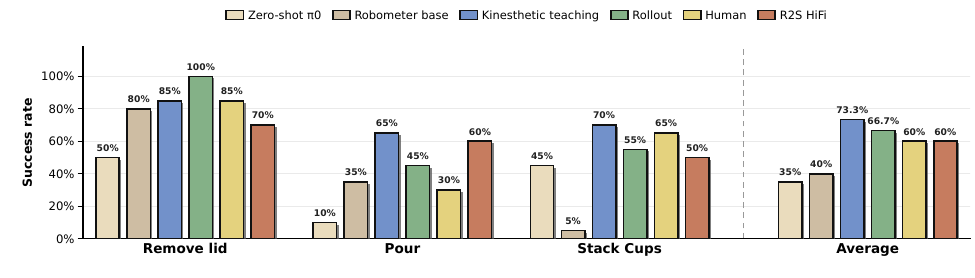}
    \caption{\textbf{Real-robot RL results (RQ1).} Success rate over 20 trials for zero-shot $\pi_0$ and DSRL with the base Robometer or source-adapted Robometer on three tasks. On average, every adapted reward improves over the base reward. Cheaper sources such as human video and real2sim are competitive with kinesthetic teaching despite their domain gaps.}
    \label{fig:DSRL_Results}
\end{figure}

\subsection{Reward models}
\label{sec:rewards}
We study two ways of using counterfactual supervision. We fine-tune Robometer-4B~\citep{liang2026robometer} with LoRA on data pooled across all five tasks. We also use Generative Value Learning (GVL)~\citep{ma2025vision}, instantiated with GPT-4o, and provide task-specific trajectories and their whole-trajectory partial-credit labels as few-shot context. Unlike TOPReward~\citep{chen2026topreward}, which reads progress from the probability of a completion token, GVL outputs progress values directly, which makes it a natural fit for few-shot prompting with our partial-credit labels. We keep the successful robot demonstrations fixed across source comparisons, so that the main difference is the failure supervision each source provides.
 
For each model, we keep the adaptation procedure fixed across counterfactual sources. We include both models in the matched-data comparison to test whether source choice matters for both parameter adaptation and in-context prediction. We use Robometer for the scaling, simulation-fidelity, and downstream RL experiments, since our GVL setting accommodates only a small number of examples in its prompt. Robometer training hyperparameters and GVL prompting details are given in Appendix~\ref{app:reward_hyperparameters} and Appendix~\ref{app:gvl}, respectively.
 
\subsection{Real-world RL setup}
\label{sec:rl_setup}
 
We use real-robot RL to test whether the choice of counterfactual supervision changes the behavior learned by a policy. We post-train the DROID checkpoint of $\pi_0$ with DSRL~\citep{wagenmaker2025steering}, which learns to steer the latent noise of the frozen base policy. After each episode, we relabel rewards with the selected reward model, combining its predicted progress with a sparse human success label (Appendix~\ref{app:rl_hyperparameters}).
 
We select three tasks spanning a range of difficulty for the RL experiments from the task pool shown in Figure~\ref{fig:teaser}: \emph{Remove Lid} (easy: a single contact-and-lift), \emph{Pour} (medium: requires controlled tilting), and \emph{Stack Cups} (hard: demands precise alignment and release). We keep the base policy, warm-up rollouts, interaction budget, and optimization hyperparameters fixed across reward sources. 

We compare zero-shot $\pi_0$ with DSRL using the base Robometer and Robometer adapted with each of the four counterfactual sources. For each source, we use the reward checkpoint at its best data scale from the scaling experiments (Section~\ref{sec:scaling}): ten trajectories per task for policy rollouts, human video, and real2sim, and five for kinesthetic teaching. We evaluate each condition in 20 blind, randomized trials with matched initial conditions and measure task success directly from the physical execution. The selected reward checkpoints and RL hyperparameters are listed in Appendix~\ref{app:rl_hyperparameters}.
 
\subsection{Offline evaluation benchmark and metrics}
\label{sec:offline_eval}
 
We use a custom offline benchmark to complement the RL experiments, since it is impractical to run online RL for every reward-model variant across all control and scaling conditions. Offline metrics are computed on a held-out benchmark of $5 \times 50 = 250$ annotated episodes: for each task, 25 zero-shot rollouts of $\pi_{0.5}$~\citep{intelligence2025pi05} and 25 of MolmoAct2~\citep{fang2026molmoact2}, each with human-assigned partial-credit labels. No $\pi_0$ rollout appears in the benchmark, so every offline evaluation is cross-policy: counterfactuals come from $\pi_0$, an operator, a human, or real2sim, while the reward model is evaluated on the behaviors of two other policies. No benchmark episode is used for fine-tuning, although we use the benchmark to select the data scale of the RL reward checkpoints (Section~\ref{sec:rl_setup}).
 
We report two primary offline metrics. \textbf{Trajectory ranking} ($\tau_{\mathrm{return}}$) measures how well the reward model orders whole trajectories by their mean predicted progress, computed as the Kendall rank correlation between the human-assigned partial credits and the trajectory-averaged reward, averaged across tasks. \textbf{Success detection} ($\text{ROC-AUC}_{\mathrm{return}}$) measures binary success classification, treating the trajectory-averaged predicted reward as a classifier score, with success defined as partial credit $1.0$. Both metrics use the trajectory-averaged prediction rather than the final-frame prediction, since trajectory-level scores better reflect how RL integrates dense per-step rewards over an episode. Frame sampling and bootstrapping details are given in Appendix~\ref{app:reward_hyperparameters}.

 \begin{figure}[!t]
    \centering
    \includegraphics[width=\linewidth]{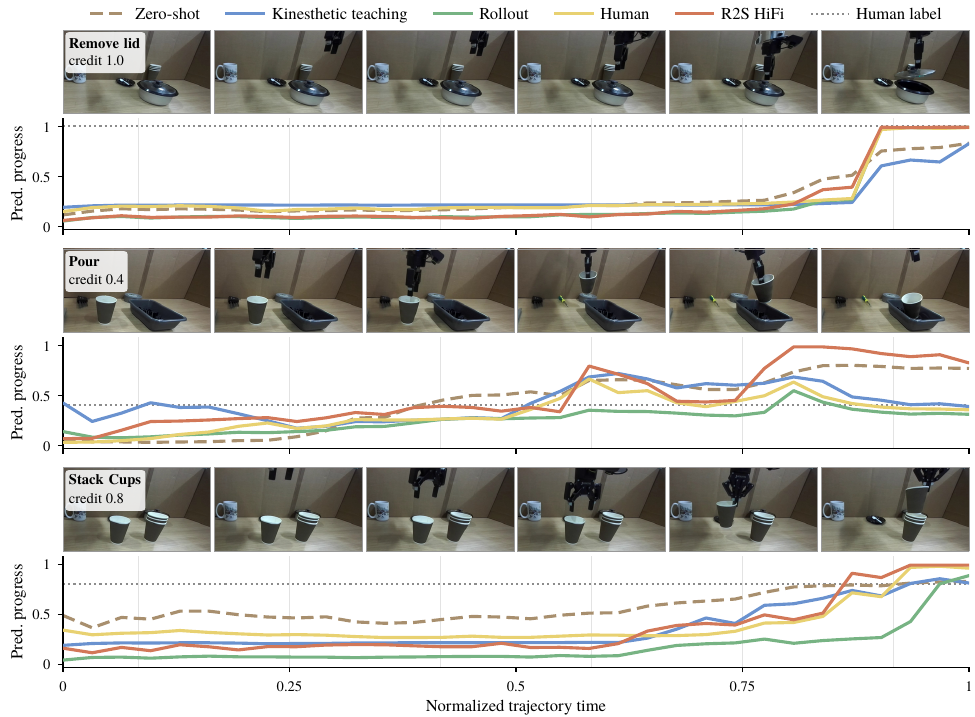}
    \vspace{-14pt}
    \caption{\footnotesize \textbf{Reward predictions on the three RL tasks.} Base and source-adapted Robometer predictions on one benchmark episode per task, with frames placed above the time they show; the dotted line is the human-assigned credit. \textbf{Remove Lid:} every model jumps when the lid comes off. \textbf{Pour} (failed, credit 0.4): the base Robometer climbs to 0.77 from the cup motion alone. \textbf{Stack Cups:} the base Robometer predicts around 0.5 before any cup moves. More examples are in Appendix~\ref{app:qualitative_results}.}
    \label{fig:qual_rl}
\end{figure}

\section{Results}
\label{sec:results}
 
We curate our results around five research questions. We first ask whether the source of counterfactuals affects how RL performs (RQ1). We then use the offline benchmark to study the sources at a fixed scale (RQ2) and when scaled according to budget (RQ3), followed by simulation fidelity (RQ4) and how well the offline metrics agree with RL (RQ5).
 
\begin{figure}[!t]
    \centering
    \includegraphics[width=\linewidth]{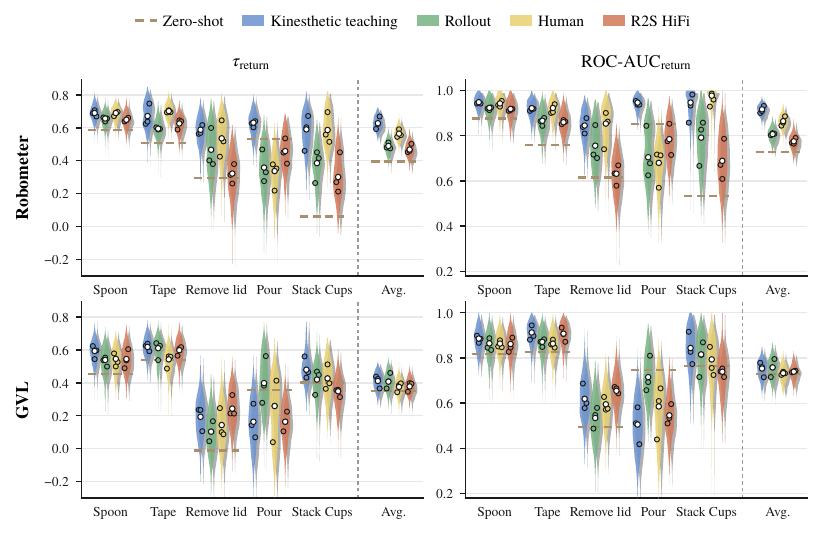}
    \caption{\textbf{Matched-scale comparison of counterfactual sources (RQ2).} Robometer (top) and few-shot GVL (bottom) adapted with five trajectories per task from each source and evaluated on the held-out benchmark, using trajectory ranking ($\tau_{\mathrm{return}}$, left) and success detection ($\text{ROC-AUC}_{\mathrm{return}}$, right). All sources improve Robometer kinesthetic teaching helps most, and human video outperforms policy rollouts. GVL barely responds to any source.}
    \label{fig:control_mean}
\end{figure}

\subsection{Real-world RL}
\label{sec:online}
 
\paragraph{RQ1: Can rewards learned from cheap counterfactual sources still give gains in RL?}\label{sec:real_rl}
Real-world RL is the most direct test for reward models and also the most expensive one, so we run it on three tasks of varying difficulty (Section~\ref{sec:rl_setup}). For each task, we compare the zero-shot policy, DSRL with the base Robometer, and DSRL with Robometer adapted to each source over 20 trials (Figure~\ref{fig:DSRL_Results}).
 
\textbf{All the counterfactual sources improve RL, including cheap sources with large domain gaps.} Averaged over the three tasks, every adapted reward improves over the base reward, from 40\% to between 60\% and 73.3\% (Figure~\ref{fig:DSRL_Results}, right). Kinesthetic teaching, the costliest source, gives the highest average, but surprisingly the cheaper sources perform competitively despite the embodiment gap in human video and the visual domain gap in real2sim. Kinesthetic teaching is the most reliable source, but its advantage is minuscule compared to its collection cost.
 
To better understand the individual results, we look at a few reward predictions from the offline benchmark (Figure~\ref{fig:qual_rl}). On Remove Lid, every model jumps when the lid comes off, which matches the RL results staying close to the base reward (70\% to 100\% vs.\ 80\%). On Stack Cups, the base Robometer rewards the scene rather than the task: the already stacked cups make it predict around 0.5 progress from the start, which drops RL below the zero-shot policy (45\% to 5\%). Pour is less clear-cut. The base Robometer climbs to 0.77 on a failed episode from the cup motion alone, while kinesthetic teaching ends close to the label. Real2sim stays close to the base here, yet gives one of the larger RL gains (60\% vs.\ 35\%), a gap we return to in RQ5 (Section~\ref{sec:correlation}).
 
\takeaway{Every source improves RL over the base reward, and the cheaper sources are competitive with kinesthetic teaching despite their domain gaps.}
 
\subsection{Offline evaluation}
\label{sec:offline}
 
Since real-robot RL is too expensive to run for every condition, we use the offline benchmark (Section~\ref{sec:offline_eval}) to measure how source, scale, and simulation fidelity change the reward itself, and whether these metrics agree with RL (RQ5).
 
\paragraph{RQ2: Which sources teach the reward model most at fixed scale?}\label{sec:control}
Holding the data fixed at five trajectories per task isolates what each source contributes. All four sources help Robometer (Figure~\ref{fig:control_mean}, top). Kinesthetic teaching gives the largest gains, but human video ranks trajectories better than policy rollouts, even with the embodiment gap. This suggests that cheap sources still carry useful failure signal despite their domain gaps, which may explain why they are competitive in RL. GVL, which only sees the counterfactuals as few-shot examples (Section~\ref{sec:rewards}), barely changes with any source (Figure~\ref{fig:control_mean}, bottom).
 
\takeaway{Kinesthetic teaching based coutnerfactuals improve Robometer the most, with human video second and ahead of policy rollouts. GVL barely changes with any counterfactual source.}
 
\begin{figure}[!t]
    \centering
    \includegraphics[width=\linewidth]{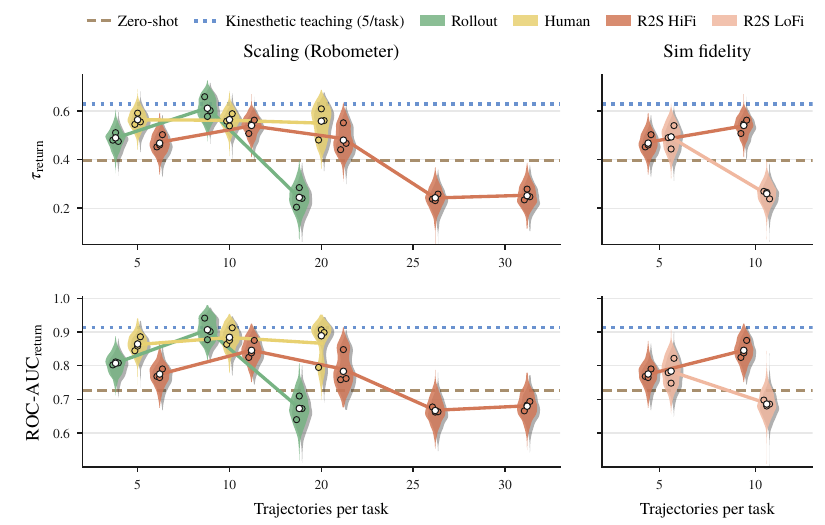}
    \caption{\textbf{Scaling and simulation fidelity (RQ3, RQ4).} \textbf{Left}: Robometer adapted with increasing cheap counterfactual data; Kinesthetic teaching at five per task (dotted) and base Robometer (dashed) for reference. Cheap sources help up to ~10 trajectories per task and can hurt beyond. \textbf{Right}: high-fidelity (R2S-HiFi) vs low-fidelity (R2S-LoFi) simulation. At small scale the two are equivalent at ten per task, only R2S-HiFi improves, showing that visual realism matters when scaling simulated data.}
    \label{fig:scaling_mean}
\end{figure}
 
\paragraph{RQ3: Is more counterfactual data always better?}\label{sec:scaling}
We scale each source according to its collection budget: kinesthetic teaching stays at five trajectories per task, policy rollouts and human video go from 5 to 10 to 20, and real2sim, the cheapest, goes up to 30 (Figure~\ref{fig:scaling_mean}, left). Human video barely changes with scale, while real2sim improves up to ten and then degrades, falling below the base Robometer at 25--30. Policy rollouts show the sharpest effect: ten per task are as good as five kinesthetic-teaching trajectories, but twenty are worse than no adaptation at all. Since training length grows with the amount of data (Appendix~\ref{app:reward_hyperparameters}), we cannot yet separate more data from longer training. Either way, cheap sources need to be scaled carefully, which is why our RL experiments use the best scale for each source (Section~\ref{sec:rl_setup}).
 
\takeaway{More data is not always better. Cheap sources help up to about ten trajectories per task and can make the reward modeling worse beyond that.}
 
\paragraph{RQ4: How real does a simulated failure need to look?}\label{sec:sim_abaltion}
How much realism a simulated failure needs depends on how much simulated data is used. At five trajectories per task, textureless MuJoCo renderings (R2S-LoFi) are as useful as Gaussian-splat renderings of the real scene (R2S-HiFi) on both metrics (Figure~\ref{fig:scaling_mean}, right). Realism starts to pay off when scaling: at ten per task, R2S-HiFi improves while R2S-LoFi drops below the base Robometer. As the two also use different simulators, this compares the full pipelines rather than rendering alone (Appendix~\ref{app:real2sim}).
 
\takeaway{Low-fidelity simulation is enough at small scale, but realism pays off once simulated data is scaled.}

\begin{figure}[!t]
    \centering
    \includegraphics[width=\linewidth]{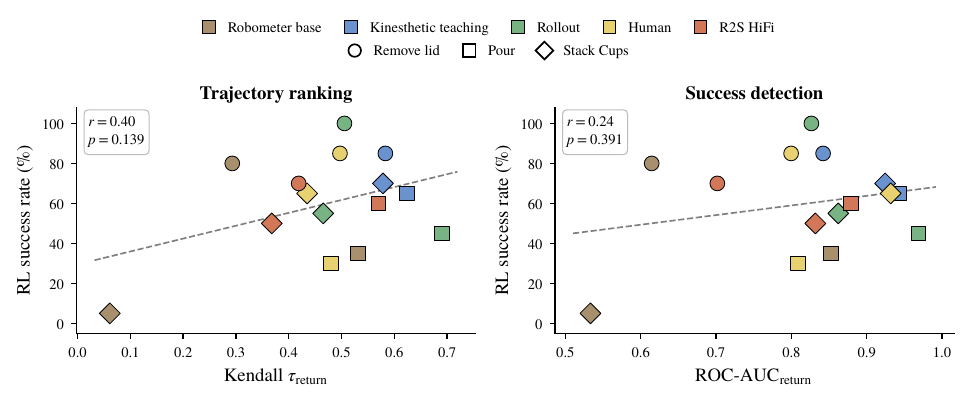}
    \caption{\textbf{Do offline metrics predict RL success? (RQ5)} Offline score vs. RL success rate for 15 (reward, task) pairs. $\tau_{\mathrm{return}}$ (r=0.40, p=0.139) tracks RL success slightly better than $\text{ROC-AUC}_{\mathrm{return}}$ (r=0.24, p=0.391).}
    \label{fig:correlation}
\end{figure}

\paragraph{RQ5: Can offline evaluation tell us which reward to deploy?}\label{sec:correlation}
We compare the offline metrics with the RL results for the 15 reward and task pairs we ran on the robot (Figure~\ref{fig:correlation}). Trajectory ranking shows moderate correaltion in comparison to success detection ($r = 0.40$, $p = 0.139$ for $\tau_{\mathrm{return}}$; $r = 0.24$, $p = 0.391$ for $\text{ROC-AUC}_{\mathrm{return}}$). The only reward that is poor on both axes is the base Robometer on Stack Cups, so the offline metrics can catch a broken reward, but a better offline score does not reliably mean a better reward for RL. Closing this gap needs better offline evaluations that reflect how a reward is actually used during RL.
 
\takeaway{Offline metrics catch a broken reward but do not reliably rank good ones. Better offline evaluation is needed.}

\section{Discussion, Limitations, and Conclusion}

Improving robot policies with RL requires rewards that recognize both progress and failure. We studied how the source and scale of counterfactual data affect these rewards and the policies trained with them. Kinesthetic-teaching failures provide strong supervision but require substantial robot and operator time. We showed that cheaper human-video and counterfactuals generated in simulation through real2sim techniques approached their performance with more data, although further scaling did not consistently improve rewards.

These conclusions come with caveats. Our trainable reward model is Robometer throughout, which already has good priors; training a reward model from scratch or adapting other VLM-based architectures may shift the source rankings, and we leave this exploration to future work. We cover five tabletop tasks on one embodiment, with RL on three of them---more tasks and embodiments would make the picture more complete. Our offline benchmark uses discrete partial credits for computing metrics, but a dense human-labeled reward curve would give more information; the cost of collecting such annotations at scale is impractical.

Our results show that a small budget of counterfactual data combined with strong priors can improve the reward regardless of the source, and cheaper sources, when scaled carefully, can match expensive ones.

\bibliography{references}

@inproceedings{radford2021learning,
  title={Learning transferable visual models from natural language supervision},
  author={Radford, Alec and Kim, Jong Wook and Hallacy, Chris and Ramesh, Aditya and Goh, Gabriel and Agarwal, Sandhini and Sastry, Girish and Askell, Amanda and Mishkin, Pamela and Clark, Jack and others},
  booktitle={International conference on machine learning},
  pages={8748--8763},
  year={2021},
  organization={PmLR}
}

@article{yang2026rarm,
  title={RARM: Confidence-Gated Progress Reward Modeling for RL in Manipulation},
  author={Yang, Pengzhi and Wang, Xinyu and Jing, Pengyu and Wen, Kehan and Qu, Yiduo and Huang, Zhenhao and Fu, Minghao and Liu, Xin and Shen, Yaheng and Shi, Fan},
  journal={arXiv preprint arXiv:2606.22027},
  year={2026}
}

@article{rajaraman2020toward,
  title={Toward the fundamental limits of imitation learning},
  author={Rajaraman, Nived and Yang, Lin and Jiao, Jiantao and Ramchandran, Kannan},
  journal={Advances in Neural Information Processing Systems},
  volume={33},
  pages={2914--2924},
  year={2020}
}

@article{zhang2025rewind,
  title={{ReWiND}: Language-guided rewards teach robot policies without new demonstrations},
  author={Zhang, Jiahui and Luo, Yusen and Anwar, Abrar and Sontakke, Sumedh Anand and Lim, Joseph J and Thomason, Jesse and Biyik, Erdem and Zhang, Jesse},
  journal={arXiv preprint arXiv:2505.10911},
  year={2025}
}

@article{lee2026roboreward,
  title={{RoboReward}: General-purpose vision-language reward models for robotics},
  author={Lee, Tony and Wagenmaker, Andrew and Pertsch, Karl and Liang, Percy and Levine, Sergey and Finn, Chelsea},
  journal={arXiv preprint arXiv:2601.00675},
  year={2026}
}

@article{tan2025robo,
  title={{Robo-Dopamine}: General process reward modeling for high-precision robotic manipulation},
  author={Tan, Huajie and Chen, Sixiang and Xu, Yijie and Wang, Zixiao and Ji, Yuheng and Chi, Cheng and Lyu, Yaoxu and Zhao, Zhongxia and Chen, Xiansheng and Co, Peterson and others},
  journal={arXiv preprint arXiv:2512.23703},
  year={2025}
}

@article{zhai2025vision,
  title={A vision-language-action-critic model for robotic real-world reinforcement learning},
  author={Zhai, Shaopeng and Zhang, Qi and Zhang, Tianyi and Huang, Fuxian and Zhang, Haoran and Zhou, Ming and Zhang, Shengzhe and Liu, Litao and Lin, Sixu and Pang, Jiangmiao},
  journal={arXiv preprint arXiv:2509.15937},
  year={2025}
}

@article{liang2026robometer,
  title={{Robometer}: Scaling general-purpose robotic reward models via trajectory comparisons},
  author={Liang, Anthony and Korkmaz, Yigit and Zhang, Jiahui and Hwang, Minyoung and Anwar, Abrar and Kaushik, Sidhant and Shah, Aditya and Huang, Alex S and Zettlemoyer, Luke and Fox, Dieter and others},
  journal={arXiv preprint arXiv:2603.02115},
  year={2026}
}

@article{lin2025failsafe,
  title={{FailSafe}: Reasoning and recovery from failures in vision-language-action models},
  author={Lin, Zijun and Duan, Jiafei and Fang, Haoquan and Fox, Dieter and Krishna, Ranjay and Tan, Cheston and Wen, Bihan},
  journal={arXiv preprint arXiv:2510.01642},
  year={2025}
}

@article{zhou2024autonomous,
  title={Autonomous improvement of instruction following skills via foundation models},
  author={Zhou, Zhiyuan and Atreya, Pranav and Lee, Abraham and Walke, Homer and Mees, Oier and Levine, Sergey},
  journal={arXiv preprint arXiv:2407.20635},
  year={2024}
}

@inproceedings{dai2025racer,
  title={{RACER}: Rich language-guided failure recovery policies for imitation learning},
  author={Dai, Yinpei and Lee, Jayjun and Fazeli, Nima and Chai, Joyce},
  booktitle={2025 IEEE International Conference on Robotics and Automation (ICRA)},
  pages={15657--15664},
  year={2025},
  organization={IEEE}
}

@article{ma2022vip,
  title={{VIP}: Towards universal visual reward and representation via value-implicit pre-training},
  author={Ma, Yecheng Jason and Sodhani, Shagun and Jayaraman, Dinesh and Bastani, Osbert and Kumar, Vikash and Zhang, Amy},
  journal={arXiv preprint arXiv:2210.00030},
  year={2022}
}

@inproceedings{ma2023liv,
  title={{LIV}: Language-image representations and rewards for robotic control},
  author={Ma, Yecheng Jason and Kumar, Vikash and Zhang, Amy and Bastani, Osbert and Jayaraman, Dinesh},
  booktitle={International Conference on Machine Learning},
  pages={23301--23320},
  year={2023},
  organization={PMLR}
}

@article{chen2026topreward,
  title={{TOPReward}: Token probabilities as hidden zero-shot rewards for robotics},
  author={Chen, Shirui and Harrison, Cole and Lee, Ying-Chun and Yang, Angela Jin and Ren, Zhongzheng and Ratliff, Lillian J and Duan, Jiafei and Fox, Dieter and Krishna, Ranjay},
  journal={arXiv preprint arXiv:2602.19313},
  year={2026}
}

@article{oquab2023dinov2,
  title={{DINOv2}: Learning robust visual features without supervision},
  author={Oquab, Maxime and Darcet, Timoth{\'e}e and Moutakanni, Th{\'e}o and Vo, Huy and Szafraniec, Marc and Khalidov, Vasil and Fernandez, Pierre and Haziza, Daniel and Massa, Francisco and El-Nouby, Alaaeldin and others},
  journal={arXiv preprint arXiv:2304.07193},
  year={2023}
}

@inproceedings{o2024open,
  title={Open {X-Embodiment}: Robotic learning datasets and {RT-X} models},
  author={O'Neill, Abby and Rehman, Abdul and Maddukuri, Abhiram and Gupta, Abhishek and Padalkar, Abhishek and Lee, Abraham and Pooley, Acorn and Gupta, Agrim and Mandlekar, Ajay and Jain, Ajinkya and others},
  booktitle={2024 IEEE International Conference on Robotics and Automation (ICRA)},
  pages={6892--6903},
  year={2024},
  organization={IEEE}
}

@article{liu2023libero,
  title={{LIBERO}: Benchmarking knowledge transfer for lifelong robot learning},
  author={Liu, Bo and Zhu, Yifeng and Gao, Chongkai and Feng, Yihao and Liu, Qiang and Zhu, Yuke and Stone, Peter},
  journal={Advances in Neural Information Processing Systems},
  volume={36},
  pages={44776--44791},
  year={2023}
}

@article{khazatsky2024droid,
  title={{DROID}: A large-scale in-the-wild robot manipulation dataset},
  author={Khazatsky, Alexander and Pertsch, Karl and Nair, Suraj and Balakrishna, Ashwin and Dasari, Sudeep and Karamcheti, Siddharth and Nasiriany, Soroush and Srirama, Mohan Kumar and Chen, Lawrence Yunliang and Ellis, Kirsty and others},
  journal={arXiv preprint arXiv:2403.12945},
  year={2024}
}

@article{atreya2025roboarena,
  title={{RoboArena}: Distributed real-world evaluation of generalist robot policies},
  author={Atreya, Pranav and Pertsch, Karl and Lee, Tony and Kim, Moo Jin and Jain, Arhan and Kuramshin, Artur and Eppner, Clemens and Neary, Cyrus and Hu, Edward and Ramos, Fabio and others},
  journal={arXiv preprint arXiv:2506.18123},
  year={2025}
}

@article{sontakke2023roboclip,
  title={{RoboCLIP}: One demonstration is enough to learn robot policies},
  author={Sontakke, Sumedh and Zhang, Jesse and Arnold, S{\'e}b and Pertsch, Karl and B{\i}y{\i}k, Erdem and Sadigh, Dorsa and Finn, Chelsea and Itti, Laurent},
  journal={Advances in Neural Information Processing Systems},
  volume={36},
  pages={55681--55693},
  year={2023}
}

@article{bai2025qwen3,
  title={{Qwen3-VL} technical report},
  author={Bai, Shuai and Cai, Yuxuan and Chen, Ruizhe and Chen, Keqin and Chen, Xionghui and Cheng, Zesen and Deng, Lianghao and Ding, Wei and Gao, Chang and Ge, Chunjiang and others},
  journal={arXiv preprint arXiv:2511.21631},
  year={2025}
}

@inproceedings{ziebart2008maximum,
  title={Maximum entropy inverse reinforcement learning},
  author={Ziebart, Brian D and Maas, Andrew L and Bagnell, J Andrew and Dey, Anind K and others},
  booktitle={AAAI},
  volume={8},
  pages={1433--1438},
  year={2008}
}

@inproceedings{abbeel2004apprenticeship,
  title={Apprenticeship learning via inverse reinforcement learning},
  author={Abbeel, Pieter and Ng, Andrew Y},
  booktitle={Proceedings of the Twenty-First International Conference on Machine Learning},
  pages={1},
  year={2004}
}

@inproceedings{brown2019extrapolating,
  title={Extrapolating beyond suboptimal demonstrations via inverse reinforcement learning from observations},
  author={Brown, Daniel and Goo, Wonjoon and Nagarajan, Prabhat and Niekum, Scott},
  booktitle={International Conference on Machine Learning},
  pages={783--792},
  year={2019},
  organization={PMLR}
}

@inproceedings{ross2011reduction,
  title={A reduction of imitation learning and structured prediction to no-regret online learning},
  author={Ross, St{\'e}phane and Gordon, Geoffrey and Bagnell, Drew},
  booktitle={Proceedings of the Fourteenth International Conference on Artificial Intelligence and Statistics},
  pages={627--635},
  year={2011},
  organization={JMLR Workshop and Conference Proceedings}
}

@inproceedings{ross2010efficient,
  title={Efficient reductions for imitation learning},
  author={Ross, St{\'e}phane and Bagnell, Drew},
  booktitle={Proceedings of the Thirteenth International Conference on Artificial Intelligence and Statistics},
  pages={661--668},
  year={2010},
  organization={JMLR Workshop and Conference Proceedings}
}

@article{wagenmaker2025steering,
  title={Steering your diffusion policy with latent space reinforcement learning},
  author={Wagenmaker, Andrew and Nakamoto, Mitsuhiko and Zhang, Yunchu and Park, Seohong and Yagoub, Waleed and Nagabandi, Anusha and Gupta, Abhishek and Levine, Sergey},
  journal={arXiv preprint arXiv:2506.15799},
  year={2025}
}

@article{black2024pi0,
  title={$\pi_0$: A vision-language-action flow model for general robot control},
  author={Black, Kevin and Brown, Noah and Driess, Danny and Esmail, Adnan and Equi, Michael and Finn, Chelsea and Fusai, Niccolo and Groom, Lachy and Hausman, Karol and Ichter, Brian and others},
  journal={arXiv preprint arXiv:2410.24164},
  year={2024}
}

@article{intelligence2025pi05,
  title={$\pi_{0.5}$: A vision-language-action model with open-world generalization},
  author={{Physical Intelligence} and Black, Kevin and Brown, Noah and Darpinian, James and Dhabalia, Karan and Driess, Danny and Esmail, Adnan and Equi, Michael and Finn, Chelsea and Fusai, Niccolo and others},
  journal={arXiv preprint arXiv:2504.16054},
  year={2025}
}

@article{torne2026freeform,
  title={Freeform Preference Learning for Robotic Manipulation},
  author={Torne, Marcel and Mahajan, Anubha and Bhat, Abhijnya and Finn, Chelsea},
  journal={arXiv preprint arXiv:2606.32027},
  year={2026}
}

@inproceedings{ma2025vision,
  title={Vision language models are in-context value learners},
  author={Ma, Yecheng Jason and Hejna, Joey and Fu, Chuyuan and Shah, Dhruv and Liang, Jacky and Xu, Zhuo and Kirmani, Sean and Xu, Peng and Driess, Danny and Xiao, Ted and others},
  booktitle={International Conference on Learning Representations},
  volume={2025},
  pages={33984--34009},
  year={2025}
}

@article{ranawaka2026simfoundry,
  title={Simfoundry: Modular and automated scene generation for policy learning and evaluation},
  author={Ranawaka, Nadun and Wong, Josiah and Pai, Wei-Lin and Chu, Wei-Teng and Dai, Tianyuan and Moghani, Masoud and Yin, Hang and Jiang, Yunfan and Durbano, Wesley and Huynh, Brandon and others},
  journal={arXiv preprint arXiv:2606.28276},
  year={2026}
}

@article{chen2026agentic,
  title={Agentic Real2Sim: Physics-based World Modeling with Vision-Language Agents},
  author={Chen, Guanxiong and Xia, Qianjun and Peng, Jiawei and Zhang, Heng and Ma, Bole and Qian, Justin and Jiao, Ziyi and Zhou, Bingyang and Ye, Luoxin and Zhang, Kaifeng and others},
  journal={arXiv preprint arXiv:2607.19190},
  year={2026}
}

@article{fang2026molmoact2,
  title={Molmoact2: Action reasoning models for real-world deployment},
  author={Fang, Haoquan and Duan, Jiafei and Clay, Donovan and Wang, Sam and Liu, Shuo and Huang, Weikai and Fan, Xiang and Tsai, Wei-Chuan and Chen, Shirui and Wang, Yi Ru and others},
  journal={arXiv preprint arXiv:2605.02881},
  year={2026}
}

@article{mandlekar2023mimicgen,
  title={Mimicgen: A data generation system for scalable robot learning using human demonstrations},
  author={Mandlekar, Ajay and Nasiriany, Soroush and Wen, Bowen and Akinola, Iretiayo and Narang, Yashraj and Fan, Linxi and Zhu, Yuke and Fox, Dieter},
  journal={arXiv preprint arXiv:2310.17596},
  year={2023}
}

@article{wu2026large,
  title={Large reward models: Generalizable online robot reward generation with vision-language models},
  author={Wu, Yanru and Yuan, Weiduo and Qi, Ang and Guizilini, Vitor and Mao, Jiageng and Wang, Yue},
  journal={arXiv preprint arXiv:2603.16065},
  year={2026}
}

@article{tian2026position,
  title={Position: Good embodied reward models need bad behavior data},
  author={Tian, Ran and Wu, Yilin and Bajcsy, Andrea},
  journal={arXiv preprint arXiv:2606.01036},
  year={2026}
}

@article{damen2020epic,
  title={The epic-kitchens dataset: Collection, challenges and baselines},
  author={Damen, Dima and Doughty, Hazel and Farinella, Giovanni Maria and Fidler, Sanja and Furnari, Antonino and Kazakos, Evangelos and Moltisanti, Davide and Munro, Jonathan and Perrett, Toby and Price, Will and others},
  journal={IEEE Transactions on Pattern Analysis and Machine Intelligence},
  volume={43},
  number={11},
  pages={4125--4141},
  year={2020},
  publisher={IEEE}
}

@inproceedings{hoque2026egodex,
  title={Egodex: Learning dexterous manipulation from large-scale egocentric video},
  author={Hoque, Ryan and Huang, Peide and Yoon, David and Zhang, Jian and others},
  booktitle={International Conference on Learning Representations},
  volume={2026},
  pages={4218--4237},
  year={2026}
}

@inproceedings{nasiriany2026robocasa365,
  title={Robocasa365: A large-scale simulation framework for training and benchmarking generalist robots},
  author={Nasiriany, Soroush and Nasiriany, Sep and Maddukuri, Abhiram and Zhu, Yuke},
  booktitle={International Conference on Learning Representations},
  volume={2026},
  pages={98643--98667},
  year={2026}
}

@article{lei2025rl,
  title={Rl-100: Performant robotic manipulation with real-world reinforcement learning},
  author={Lei, Kun and Li, Huanyu and Yu, Dongjie and Wei, Zhenyu and Guo, Lingxiao and Jiang, Zhennan and Wang, Ziyu and Liang, Shiyu and Xu, Huazhe},
  journal={arXiv preprint arXiv:2510.14830},
  year={2025}
}

@article{luo2025precise,
  title={Precise and dexterous robotic manipulation via human-in-the-loop reinforcement learning},
  author={Luo, Jianlan and Xu, Charles and Wu, Jeffrey and Levine, Sergey},
  journal={Science Robotics},
  volume={10},
  number={105},
  pages={eads5033},
  year={2025},
  publisher={American Association for the Advancement of Science}
}

@article{dong2026expo,
  title={EXPO-FT: Sample-Efficient Reinforcement Learning Finetuning for Vision-Language-Action Models},
  author={Dong, Perry and Hung, Kuo-Han and Gao, Tian and Sadigh, Dorsa and Finn, Chelsea},
  journal={arXiv preprint arXiv:2605.25477},
  year={2026}
}

@inproceedings{yin2026emergent,
  title={Emergent dexterity via diverse resets and large-scale reinforcement learning},
  author={Yin, Patrick and Westenbroek, Tyler and Zhang, Zhengyu and Dagnino, Ignacio and Shilamkar, Eeshani and Mbiziwo-Tiapo, Numfor and Bagaria, Simran and Liu, Xinlei and Mullins, Galen and Kolobov, Andrey and others},
  booktitle={International Conference on Learning Representations},
  volume={2026},
  pages={88343--88356},
  year={2026}
}

@article{zhang2026balanced,
  title={A balanced data diet: Addressing the exploration bottleneck in mega-scale rl for robot control},
  author={Zhang, Octi and Castro, Mateo Guaman and Yin, Patrick and Dagnino, Ignacio and Gupta, Abhishek and Scalise, Rosario and Boots, Byron},
  journal={arXiv preprint arXiv:2610.12465},
  year={2026}
}

@article{gupta2022unpacking,
  title={Unpacking reward shaping: Understanding the benefits of reward engineering on sample complexity},
  author={Gupta, Abhishek and Pacchiano, Aldo and Zhai, Yuexiang and Kakade, Sham and Levine, Sergey},
  journal={Advances in Neural Information Processing Systems},
  volume={35},
  pages={15281--15295},
  year={2022}
}
\bibliographystyle{iclr2027_conference}

\newpage
\clearpage
\appendix
\section{Appendix}
\label{app:intro}
The appendix gives more details on how we collect and label counterfactual data (Appendix~\ref{app:counterfactual_sources}), how we train and prompt the reward models (Appendix~\ref{app:reward_hyperparameters}), and how we run RL (Appendix~\ref{app:rl_hyperparameters}). Our robot is set up with the \texttt{droid-fr3} codebase\footnote{\url{https://github.com/EmbodiedForge/droid-fr3}}, the reward models are trained with the Robometer codebase\footnote{\url{https://github.com/robometer/robometer}}, and RL is run with the Robometer policy-learning codebase\footnote{\url{https://github.com/robometer/robometer-policy-learning}}.
\subsection{Counterfactual sources and annotation}
\label{app:counterfactual_sources}
 
To make data easy to inspect for all four sources, with their partial-credit labels, can be explored in our \href{https://huggingface.co/spaces/anonymouslukog/robometer-data-explorer}{\textbf{counterfactual data explorer}}. It has kinesthetic-teaching failures, policy rollouts, human video, and real2sim trajectories for all five tasks.
 
\paragraph{Partial-credit annotation}
We label every trajectory with a partial credit from $\mathcal{P}=\{0.0,0.2,0.4,0.6,0.8,1.0\}$, where $0.0$ means no real progress and $1.0$ means the task was completed. Values in between say on average how the task was done. Kinesthetic-teaching, policy-rollout, and human-video trajectories are labeled by human, while real2sim get their labels automatically from the task milestones in the simulator (Appendix~\ref{app:failgen}). All four sources use the same scale.
 
\subsubsection{Real2sim scene construction}
\label{app:real2sim}

For each task, we use one successful kinesthetic demonstration to build the same scene in two simulators: OmniGibson for high fidelity (R2S-HiFi) and MuJoCo for low fidelity (R2S-LoFi). Both keep the robot, task objects, initial configurations, and task definition, but they differ in rendering and in the simulator itself.

\paragraph{High-fidelity real2sim}
The high-fidelity scene is built with parts of SimFoundry~\citep{ranawaka2026simfoundry}. From calibrated exterior and wrist camera recordings, we get the scene geometry, object poses, and camera alignment. Each object is segmented and reconstructed as a textured 3D mesh with collision geometry before being imported into OmniGibson, and a Gaussian-splat reconstruction of the workspace gives the background. We had to define the collision geometry for hollow and thin objects (the cups, grey box, tape, and pot lid) so that grasping, insertion, and pouring behave properly. Finally, the scene is aligned with the real robot using the camera calibration and robot kinematics.

\paragraph{Low-fidelity real2sim}
Instead of reconstructing the scene again, we reuse the R2S-HiFi assets. We move over the object meshes, initial poses, robot configuration, camera calibration, and task predicates, and use the same Franka arm and Robotiq gripper. The only catch is that we use textureless assets, and the background is a standard simulator background with no details, unlike R2S-HiFi. This gives a visually low-fidelity version of the exact same task, so every R2S-LoFi trajectory has a matching R2S-HiFi trajectory. Since the simulator also changes, the fidelity ablation in RQ4 (Section~\ref{sec:sim_abaltion}) compares the two full pipelines.
 
\paragraph{Successful demonstration replay}
From the successful kinesthetic demonstration, we keep the joint positions, gripper commands, end-effector poses, and camera calibration. Long idle parts are cut out, but anything that matters for the task, like contact, grasping, and release, is kept, with a short settling period wherever needed. We then replay this edited trajectory in both simulators and keep it only if it succeeds and ends in a stable state. This replay is then used to make failures with the algorithm described below (Appendix~\ref{app:failgen}).
 
\subsubsection{Real2sim failure generation}
\label{app:failgen}

We make failures by replaying the successful demonstration and perturbing it at different task stages, similar in spirit to MimicGen~\citep{mandlekar2023mimicgen} but for failures. First, we change the initial object configuration and retarget the reference trajectory relative to the manipulated object. For an end-effector pose $T_{WE}$ and object pose $T_{WO}$, the new retargeted end-effector pose is
\begin{equation}
T'_{WE} = T'_{WO}\,T_{WO}^{-1}\,T_{WE},
\end{equation}
where $W$, $O$, and $E$ are the world, object, and end-effector frames. This keeps the end-effector in the same pose relative to the object.

\begin{sloppypar}
Each failure comes from one operator applied to one task stage. The operators are \texttt{truncate\_before}, \texttt{no\_grip}, \texttt{collision\_nudge}, \texttt{under\_actuate}, \texttt{halt\_mid\_transport}, \texttt{early\_release}, and \texttt{goal\_offset}. These operators range over a set of commonly observed, realistic failures. The part of the trajectory before the chosen stage is kept as it is, and the edited part is solved with inverse kinematics and interpolated (Algorithm~\ref{alg:failgen}).
\end{sloppypar}
 
\paragraph{Automatic labels}
Each task is split into roughly five ordered milestones (contact with the object, a secure grasp, lifting or partial manipulation, alignment or placement, and task completion), detected from the simulator states. The partial credit is assigned using
\begin{equation}
p = 0.2k, \qquad k\in\{0,1,2,3,4,5\},
\end{equation}
where $k$ is the number of milestones reached in order. A milestone counts only if the ones before it were reached. A failure is kept only if the labels are agreed upon in both simulators and the final state is stable.

\begin{tcolorbox}[colback=sim!6, colframe=simtext, boxrule=0.6pt, arc=6pt,
  left=6pt, right=6pt, top=4pt, bottom=4pt, before skip=8pt, after skip=10pt,
  title={\refstepcounter{algorithm}\label{alg:failgen}\textbf{Algorithm~\thealgorithm: Real2sim counterfactual generation}},
  fonttitle=\normalsize, coltitle=white, colbacktitle=simtext]
\begin{algorithmic}[1]
\Require Successful demonstration $\tau$; task milestones $\mathcal{M}$; failure operators $\mathcal{O}$; target failure count $N_{\mathrm{fail}}=29$
\Ensure Selected failures $\mathcal{D}^{-}$ and successful replay $\tau^{+}$

\Statex \textcolor{simtext}{\footnotesize\textit{// Phase 1: Prepare a validated successful replay}}
\State $\tau^{+}\gets\textsc{CompressIdle}(\tau)$
\State Preserve contact, grasp, release, and task-critical windows
\State Replay $\tau^{+}$ in MuJoCo and OmniGibson
\State Verify ordered milestones, task success, and stable final state

\Statex
\Statex \textcolor{simtext}{\footnotesize\textit{// Phase 2: Generate candidate failures}}
\State $\mathcal{C}\gets\emptyset$
\While{insufficient validated candidate coverage}
    \State Sample matched initial object-state perturbation
    \State Retarget the reference trajectory in the object frame
    \State Sample failure stage, operator $o\sim\mathcal{O}$, and operator parameters
    \State Modify the selected stage while preserving the valid prefix
    \State Solve inverse kinematics and interpolate commands
    \State $\tau_{\mathrm{MJ}}\gets\textsc{ReplayMuJoCo}(\tilde{\tau})$
    \State $\tau_{\mathrm{OG}}\gets\textsc{ReplayOmniGibson}(\tilde{\tau})$
    \State Compute the longest ordered milestone prefix in each simulator
    \State Assign $p=0.2k$
    \If{both simulators fail, agree on $p$, preserve milestone order, and terminate stably}
        \State Add the candidate, reward, states, and videos to $\mathcal{C}$
    \EndIf
\EndWhile

\Statex
\Statex \textcolor{simtext}{\footnotesize\textit{// Phase 3: Select and export the dataset}}
\State Cover different failure-operator outcomes and partial-credit levels
\State Select $N_{\mathrm{fail}}=29$ failures using visual-diversity sampling
\State $\mathcal{D}^{-}\gets\textsc{SelectFailures}(\mathcal{C},29)$
\State Export matched R2S-HiFi/R2S-LoFi videos, metadata, and rewards
\State \Return $\mathcal{D}^{-},\tau^{+}$
\end{algorithmic}
\end{tcolorbox}

Out of all the saved failures, we select 29 failures per task. The selection makes sure that diverse failure types and credit levels are covered, using visual-diversity sampling. Together with the successful replay, this gives 30 episodes per task in both R2S-HiFi and R2S-LoFi. Human effort is only needed once per task, to collect the successful demo and set up the scene and milestones. Everything after that runs automatically.
 
\subsection{Reward model training}
\label{app:reward_hyperparameters}
For a fair comparison, we keep the training setup for Robometer and the prompting setup for GVL fixed across all sources. Only the counterfactual data source and its scale change.

\subsubsection{Robometer}

We start from the released \texttt{Robometer-4B} checkpoint, which uses Qwen3-VL-4B-Instruct as its backbone. We keep the vision encoder frozen and only adapt the language model with LoRA, applied to all attention and MLP projections. Robometer has three heads that are trained together: preference, progress, and success. Each head is a small MLP; the preference and success heads output a single logit, and the progress head outputs logits for ten progress bins. The full configuration is in Table~\ref{tab:app_robometer_hyperparameters}.
 
\begin{table}[t]
\centering
\caption{\textbf{Robometer adaptation hyperparameters.} The same configuration is used for every counterfactual source.}
\label{tab:app_robometer_hyperparameters}
\small
\setlength{\tabcolsep}{5pt}
\renewcommand{\arraystretch}{1.08}
\begin{tabularx}{\linewidth}{@{}lX@{}}
\toprule
Parameter & Configuration \\
\midrule
Starting checkpoint & Robometer-4B \\
Backbone & Qwen3-VL-4B-Instruct \\
Trainable components & Language-model LoRA + three Robometer heads \\
Vision encoder & Frozen \\
LoRA rank / alpha / dropout & $16$ / $32$ / $0.05$ \\
Optimizer & AdamW \\
Learning rate & $1\times10^{-4}$ \\
Weight decay & $0.01$ \\
Scheduler / warm-up & Cosine / $10\%$ of training steps \\
Adam $(\beta_1,\beta_2,\epsilon)$ & $(0.9,0.999,10^{-8})$ \\
Maximum gradient norm & $1.0$ \\
Batch size / accumulation & $2$ / $1$ \\
Precision & BF16 \\
Gradient checkpointing & Enabled \\
Maximum sequence length & $1024$ \\
Training seeds & $42,\ 43,\ 44$ \\
Training steps & $3N$, where $N$ is the number of training trajectories \\
Maximum frames per training sample & $8$ \\
\bottomrule
\end{tabularx}
\end{table}
 
\paragraph{Sampling and supervision}
Each training data point samples up to eight frames per trajectory. Preference pairs are built from suboptimal trajectories of the same task, rewound trajectories, trajectories from different tasks, and reversed progress. For suboptimal trajectories, two partial-credit trajectories are paired only if their labels differ by at least 0.2. For a failed or partial trajectory, the label only supervises the last frame, with no progress labels in between. The rest of the progress signal comes from augmented trajectories and successful ones. The training loss is
\begin{equation}
\mathcal{L} = \mathcal{L}_{\mathrm{pref}} + \mathcal{L}_{\mathrm{progress}} + \mathcal{L}_{\mathrm{success}},
\end{equation}
where $\mathcal{L}_{\mathrm{pref}}$ is a binary cross-entropy loss over the constructed preference pairs, $\mathcal{L}_{\mathrm{progress}}$ is cross-entropy over ten bins, and $\mathcal{L}_{\mathrm{success}}$ is a masked binary cross-entropy, all weighted equally. For more details, refer to the Robometer paper~\citep{liang2026robometer}.

Training runs for $3N$ steps, where $N$ is the number of training trajectories across the five tasks, so larger datasets also train longer. Each smaller training set is a subset of the larger ones, and R2S-LoFi always uses the same episodes as R2S-HiFi for a fair visual-fidelity comparison.

\paragraph{Offline evaluation}
We evaluate every model on the 250 held-out benchmark trajectories mentioned in Section~\ref{sec:offline_eval}. Each trajectory is uniformly subsampled to 30 frames, and the reported metrics are computed with 10,000 bootstrap samples.
 
\subsubsection{GVL}
\label{app:gvl}
GVL is our frozen, in-context reward model. We query \texttt{gpt-4o} through the API, keeping the prompt and task description fixed and only changing the in-context bank for each counterfactual source. For zero-shot GVL results, we just use the default GVL setup to evaluate. For the few-shot setting, we sample five examples with their assigned partial credits from each counterfactual source. While sampling, we make sure the labels per source are well spread, giving diverse partial credits. Table~\ref{tab:app_gvl_hyperparameters} summarizes the settings.

For each trajectory, we sample up to 30 frames uniformly, and we do the same for each few-shot example trajectory, which is appended with its partial credit. The evaluation frames are sent in a shuffled order, and the model returns a JSON list with a completion percentage for every frame. The parsed output is then divided by 100, and the completion of each frame is put back in the original order.
 
\begin{table}[t]
\centering
\caption{\textbf{GVL configuration.} Seeds change the frame order, not the model.}
\label{tab:app_gvl_hyperparameters}
\small
\setlength{\tabcolsep}{5pt}
\renewcommand{\arraystretch}{1.08}
\begin{tabularx}{\linewidth}{@{}lX@{}}
\toprule
Parameter & Configuration \\
\midrule
Model & \texttt{gpt-4o} \\
Adaptation & Frozen model; zero-shot or few-shot prompting \\
Few-shot context & Five same-task trajectories per source \\
Target frames & Up to $30$, uniformly sampled \\
Context frames & Up to $30$ per example trajectory \\
Maximum response tokens & $4096$ \\
Seeds & $42,\ 43,\ 44$ \\
Output & JSON with per-frame completion percentages \\
\bottomrule
\end{tabularx}
\end{table}
 
\subsection{RL details}
\label{app:rl_hyperparameters}
For all three RL tasks (Remove Lid, Pour, and Stack Cups), we share the same policy, network, and SAC settings (Tables~\ref{tab:app_rl_observations} and~\ref{tab:app_dsrl_hyperparameters}). We only change the maximum number of control steps and the number of warm-up steps (Table~\ref{tab:app_rl_tasks}).

\subsubsection{Policy and observations}
We use \texttt{pi0-DROID} as the base policy, which stays frozen, and only train the DSRL noise-space actor and critics. We use the $\pi_0$ flow horizon of ten, while SAC predicts eight noise rows and the robot executes eight actions per transition. We copy the first two noise rows into the two unused positions. The policy uses two cameras, one exterior and one wrist, along with proprioception and the task instruction. The reward model uses only the exterior camera.

\begin{table}[t]
\centering
\caption{\textbf{Policy and observation configuration}, shared by all three RL tasks.}
\label{tab:app_rl_observations}
\small
\setlength{\tabcolsep}{5pt}
\renewcommand{\arraystretch}{1.08}
\begin{tabularx}{\linewidth}{@{}lX@{}}
\toprule
Component & Configuration \\
\midrule
Frozen base policy & \texttt{pi0-DROID} \\
Flow horizon & $10$ \\
Executed action chunk & $8$ actions \\
SAC noise action & $8\times32$, bounded to $[-2,2]$ \\
Robot-state observation & $8$ dimensions \\
Visual observation & $384$-D frozen DINOv2-small embedding \\
Policy feature & $2048$-D $\pi_0$ prefix/VLM feature \\
Reward-model input & Left exterior view only \\
\bottomrule
\end{tabularx}
\end{table}

\subsubsection{DSRL actor, critics, and SAC}
We use Robometer's adaptation~\citep{liang2026robometer} of the original DSRL setup~\citep{wagenmaker2025steering}, which changes both the actor and critic in noise space to use DINOv2~\citep{oquab2023dinov2} features with a chunking transformer architecture. Table~\ref{tab:app_dsrl_hyperparameters} lists the hyperparameters we use. For further details, refer to the Robometer policy-learning codebase (Appendix~\ref{app:intro}) and the appendix of the Robometer paper~\citep{liang2026robometer}.

\begin{table}[t]
\centering
\caption{\textbf{DSRL SAC hyperparameters}, shared by all three RL tasks.}
\label{tab:app_dsrl_hyperparameters}
\small
\setlength{\tabcolsep}{5pt}
\renewcommand{\arraystretch}{1.08}
\begin{tabularx}{\linewidth}{@{}lX@{}}
\toprule
Parameter & Configuration \\
\midrule
Transformer width / heads / layers & $256$ / $8$ / $6$ \\
Actor MLP & $[512,512]$, ReLU \\
Critic MLP & $[768,512]$, ReLU + LayerNorm \\
Critics / critics per update & $4$ / $2$ \\
Batch size & $64$ \\
Replay capacity & $10{,}000$ transitions \\
SAC train steps & $10$ per rollout transition step \\
Critic updates per actor update & $4$ \\
Actor learning rate & $5\times10^{-5}$ \\
Critic learning rate & $1\times10^{-4}$ \\
Entropy-coefficient learning rate & $1\times10^{-4}$ \\
Actor Adam $\epsilon$ / weight decay & $10^{-8}$ / $10^{-6}$ \\
Critic Adam $(\beta_1,\beta_2)$ & $(0.95,0.999)$ \\
Critic Adam $\epsilon$ / weight decay & $10^{-8}$ / $10^{-6}$ \\
Entropy coefficient & Automatic; target entropy $0$ \\
Base discount $\gamma$ & $0.995$ \\
Effective transition discount & $0.995^8 \approx 0.9607$ \\
Target-network $\tau$ & $0.005$ \\
Target update interval & $1$ \\
Precision / seed & FP16 / $0$ \\
\bottomrule
\end{tabularx}
\end{table}

\subsubsection{Warm-up and online training}
Each task has a different maximum number of control steps per rollout, all at 15\,Hz. Before online training starts, we collect a warm-up dataset for SAC by sampling uniform noise. Every reward condition uses this same warm-up from scratch, so all RL runs across the different counterfactual sources start from the same point. For the per-task max control steps, warm-up steps, and online interaction steps, refer to Table~\ref{tab:app_rl_tasks}.

\begin{table}[t]
\centering
\caption{\textbf{Per-task RL settings.} Everything else is shared across tasks (Tables~\ref{tab:app_rl_observations} and~\ref{tab:app_dsrl_hyperparameters}).}
\label{tab:app_rl_tasks}
\small
\setlength{\tabcolsep}{6pt}
\renewcommand{\arraystretch}{1.1}
\begin{tabular}{@{}lccc@{}}
\toprule
 & Remove Lid & Pour & Stack Cups \\
\midrule
Max control steps & $400$ & $500$ & $400$ \\
Warm-up steps & $2{,}000$ & $5{,}000$ & $10{,}000$ \\
Online interaction steps & $10{,}000$ & $10{,}000$ & $10{,}000$ \\
\bottomrule
\end{tabular}
\end{table}

\subsubsection{Reward and selected checkpoints}
After each episode, Robometer scores the trajectory with its progress prediction, which is then combined with the sparse human reward of $-1$ for every step and $0$ when the episode ends successfully. We use one Robometer checkpoint per source for RL, since it is trained jointly on all tasks. Since the cheap sources can be collected at several scales, we pick the scale and seed of each source that did best offline. This ends up being ten trajectories per task for policy rollouts, human video, and R2S-HiFi, while kinesthetic teaching uses five, as seen in Table~\ref{tab:app_rl_checkpoints}.

\begin{table}[t]
\centering
\caption{\textbf{Reward checkpoints used for RL.} For each source, we use the checkpoint that performs best offline; the same checkpoint per source is used for all three RL tasks.}
\label{tab:app_rl_checkpoints}
\small
\setlength{\tabcolsep}{8pt}
\renewcommand{\arraystretch}{1.1}
\begin{tabular}{@{}lc@{}}
\toprule
Source & Trajectories per task \\
\midrule
Kinesthetic & $5$ \\
Rollout & $10$ \\
Human & $10$ \\
R2S-HiFi & $10$ \\
\bottomrule
\end{tabular}
\end{table}

\subsection{Additional qualitative reward results}
\label{app:qualitative_results}
All reward curves, per task and per reward condition, can be found in our \href{https://huggingface.co/spaces/anonymouslukog/robometer-benchmark-explorer}{\textbf{reward benchmark explorer}}. It shows the predictions together with their human partial-credit labels, so the reward behavior can be compared on all benchmark trajectories beyond the examples in Figure~\ref{fig:qual_rl}.

\end{document}